\pdfoutput=1

\documentclass[11pt]{article}

\usepackage{acl}

\usepackage{times}
\usepackage{latexsym}

\usepackage[T1]{fontenc}

\usepackage[utf8]{inputenc}

\usepackage{microtype}

\usepackage{algorithm}
\usepackage{algorithmic}
\usepackage{multirow}
\usepackage{multicol}
\usepackage{booktabs}
\usepackage{marvosym}
\usepackage{colortbl}
\usepackage{arydshln}
\usepackage{pgfplots}
\usepackage{subfigure}
\usepackage{amsmath}
\usepackage{amsfonts}
\title{BORT: Back and Denoising Reconstruction \\for End-to-End Task-Oriented Dialog}

\author{ Haipeng Sun, Junwei Bao\thanks{\;\;Corresponding author.}, Youzheng Wu, Xiaodong He\\
	JD AI Research, Beijing, China \\
	\texttt{\{sunhaipeng6, 
baojunwei, wuyouzheng1, hexiaodong\}@jd.com} \\
\\}

\begin{document}
\maketitle
\begin{abstract} 

A typical end-to-end task-oriented dialog system transfers context into dialog state, and upon which generates a response, which usually faces the problem of error propagation from both previously generated inaccurate dialog states and responses, especially in low-resource scenarios.  To alleviate these issues, we propose BORT, a \textbf{b}ack and den\textbf{o}ising \textbf{r}econs\textbf{t}ruction approach for end-to-end task-oriented dialog system.  Squarely, to improve the accuracy of dialog states, \textit{back reconstruction} is used to reconstruct the original input context from the generated dialog states since  inaccurate  dialog states cannot recover the corresponding input context. To enhance the denoising capability of the model to reduce the impact of error propagation, \textit{denoising reconstruction} is used to reconstruct the corrupted dialog state and response.  Extensive experiments conducted on MultiWOZ 2.0 and CamRest676 show the effectiveness of BORT. Furthermore, BORT demonstrates its advanced capabilities in the  zero-shot domain and  low-resource scenarios\footnote{The code is available at \url{https://github.com/JD-AI-Research-NLP/BORT}.}. 
\end{abstract}

\section{Introduction}

Recently, task-oriented dialog systems, which aim to assist users to complete some booking tasks, have attracted great interest in the research community and the industry~\cite{DBLP:journals/corr/abs-2003-07490}. Task-oriented dialog systems have been usually established via a pipeline system, including several modules such as natural language understanding, dialog state tracking, dialog policy, and natural language generation.
The natural language understanding module converts user utterance into the structured semantic representation. The dialog state generated by the dialog state tracking module is used to query the database to  achieve  matched entities. The natural language generation module converts the action state estimated by the dialog policy module to the natural language response.
This  modular-based architecture is highly interpretable and easy to implement,  used in most practical task-oriented dialog systems in the industry. However, every module is  optimized individually and doesn't  consider the entire dialog history, which affects the performance of the dialog system. 
Therefore, many researchers focus on end-to-end task-oriented dialog systems to train an overall mapping from user natural language input to system natural language output \cite{lei-etal-2018-sequicity,DBLP:conf/aaai/ZhangOY20,DBLP:conf/nips/Hosseini-AslMWY20,lin-etal-2020-mintl,DBLP:conf/aaai/YangLQ21}. 

\begin{figure}[t]
  \centering
  \includegraphics[width=2.7in]{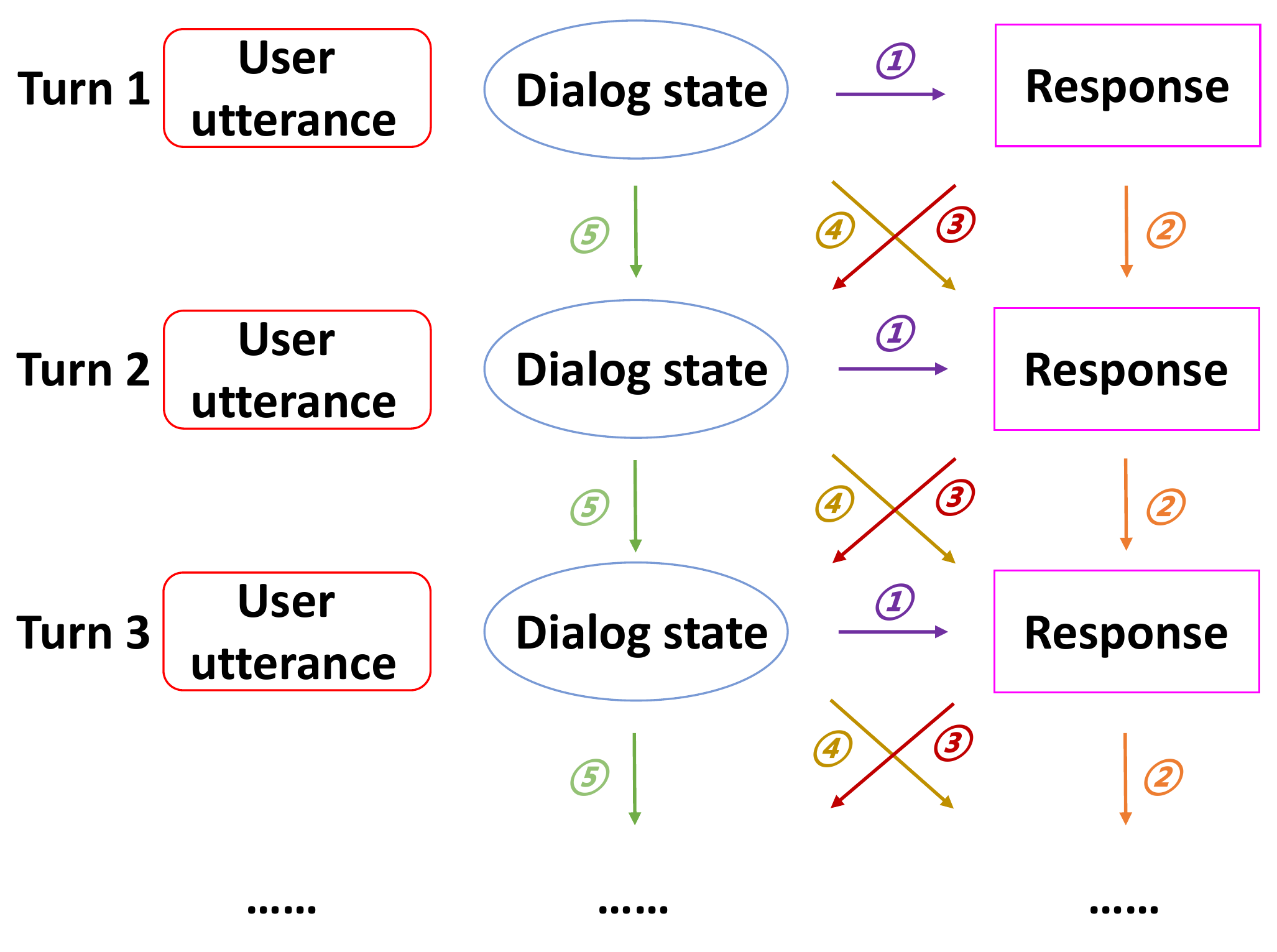}
  \caption{Illustration of different error propagation problem types, denoted by arrows in different colors, along the multi-turn task-oriented dialog flow. For example, the orange arrow indicates that the error in the previously generated response would affect the response generation in the current dialog turn.}
  \label{fig:error}
\end{figure}

However, all  existing task-oriented dialog systems still suffer from one or more types of  error propagation problems from both previously generated inaccurate dialog states and responses, as is illustrated in Figure~\ref{fig:error}.
Firstly, the generated dialog state, which is crucial for task completion of task-oriented dialog systems, is usually inaccurate across the end-to-end task-oriented dialog system training. Secondly, the previously generated  dialog state and response are encoded to create the current dialog state and response during inference, while the oracle previous dialog state and response are encoded during training. There exists a discrepancy between training and inference, affecting the quality of generated system responses. 

We propose BORT, a back and denoising reconstruction approach for end-to-end task-oriented dialog systems to alleviate these issues.
To improve dialog state learning ability, back reconstruction is used to reconstruct the generated dialog state back to the original input context to ensure that the information in the input side is completely transformed to the output side. 
In addition, to enhance the denoising capability of the task-oriented dialog system to reduce the impact of error propagation, 
denoising reconstruction is used to reconstruct  the corrupted dialog state and response. It guarantees  that the system  learns enough internal information of the dialog context to  recover the original version.
Experimental results on MultiWOZ 2.0 and CamRest676 show that our proposed BORT substantially outperforms baseline systems.
This paper primarily makes the following contributions:

\begin{itemize}
	\item We propose two effective reconstruction strategies, i.e., back and denoising reconstruction
strategies, to improve the performance of end-to-end task-oriented dialog systems.
	
	\item   Extensive experiments and analysis on MultiWOZ 2.0 and CamRest676 show the effectiveness of BORT.	
	\item   BORT  achieves promising performance in zero-shot domain scenarios and alleviates poor performance in low-resource scenarios.
\end{itemize}

\begin{figure*}[ht]
  \centering
  \subfigure{
  \begin{minipage}[b]{0.48\linewidth}
  \centering
  \includegraphics[width=3in]{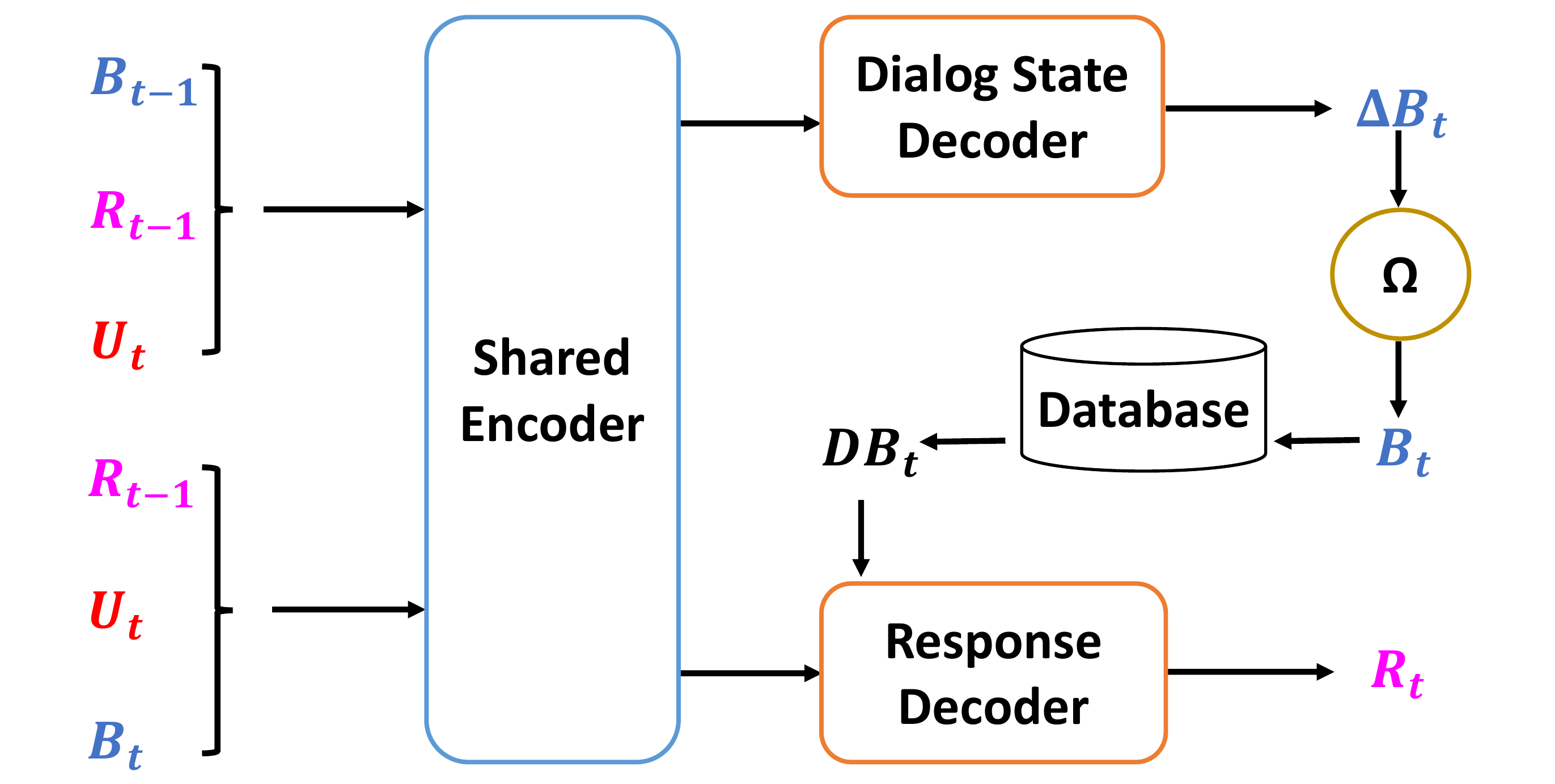}
  \centerline{(a) {Base architecture}}\label{fig:baseline}

  \end{minipage}
  }
  \subfigure{
  
  \begin{minipage}[b]{0.48\linewidth}
  \centering
  \includegraphics[width=3in]{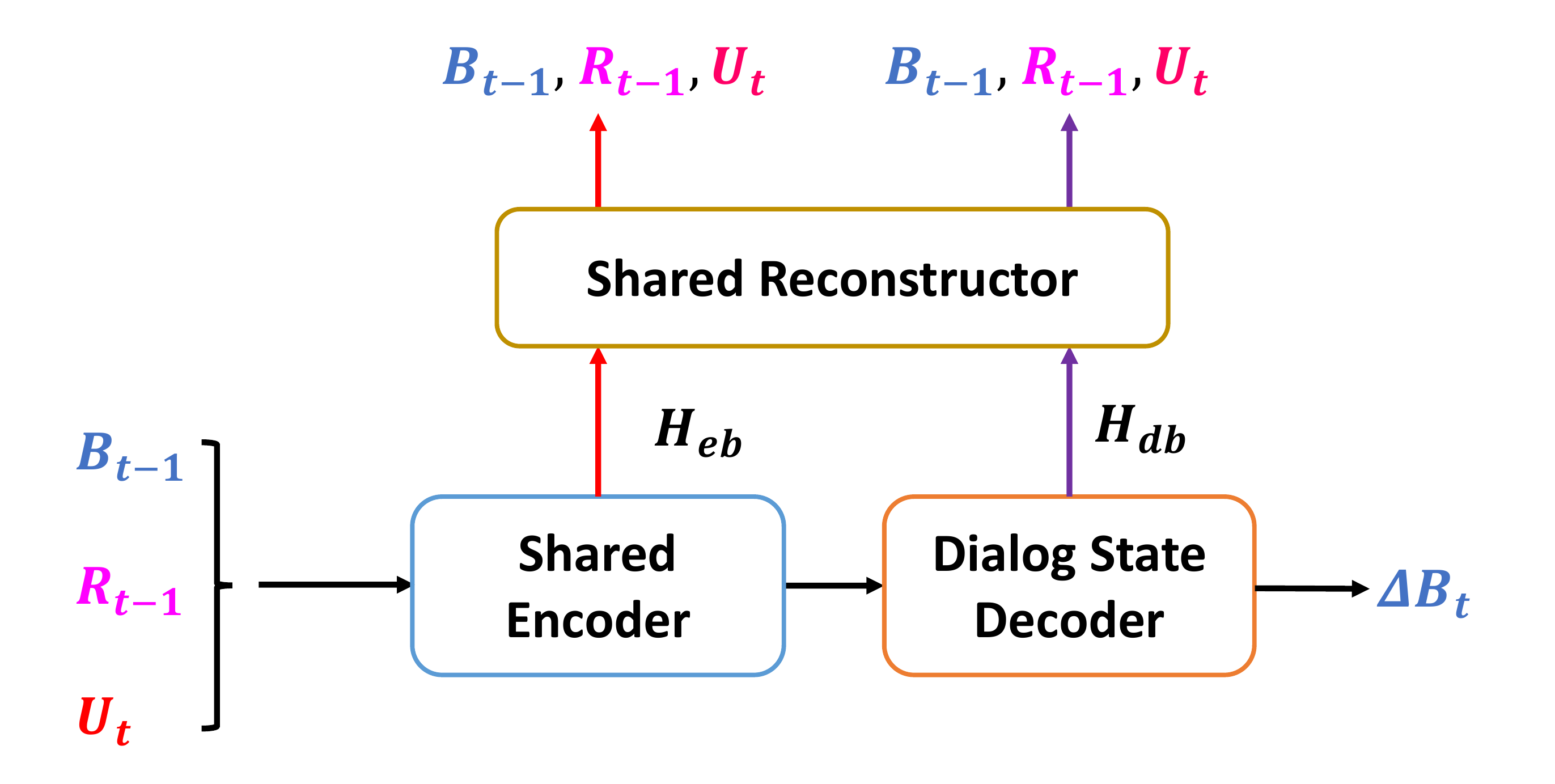}
  \centerline{(b) {Back reconstruction}}\label{fig:BR}
  \end{minipage}
  }
     \subfigure{
     \begin{minipage}[b]{0.96\linewidth}
  \centering
  \includegraphics[width=6in]{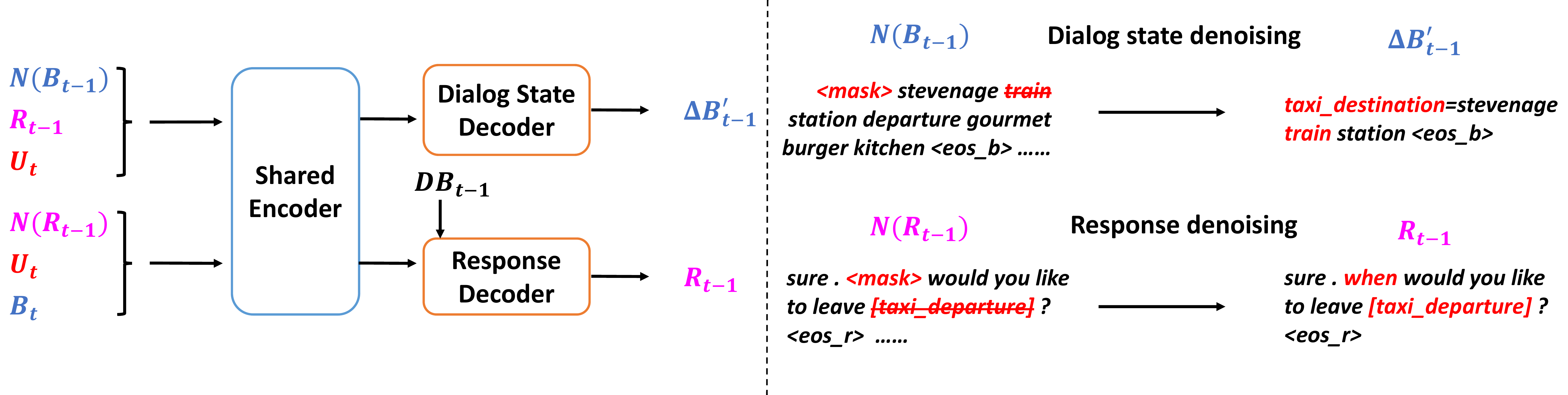}
    \centerline{(c) {Denoising reconstruction} }\label{fig:dr}
  \end{minipage}
  }

  \caption{Illustration of the task-oriented dialog  training process. We take turn \textit{t} of a dialog session as an example. }
  \label{fig:architecture}
\end{figure*}

\section{Task-Oriented Dialog Framework}
As illustrated in Figure \ref{fig:baseline}, we construct an encoder-decoder framework for an end-to-end task-oriented dialog  system via dialog state tracking and response generation tasks. 
One shared encoder encodes dialog context, and two different decoders  decode dialog state and system response, respectively.
The objective function $\mathcal{L}_{all}$ of the entire training process is optimized as:

{\footnotesize\begin{equation}
\begin{aligned}
\mathcal{L}_{all} = \mathcal{L}_{B} + \mathcal{L}_{R},
\end{aligned}
\end{equation}}%
where $ \mathcal{L}_{B}$ is the objective function for dialog state tracking, and $ \mathcal{L}_{R}$ is the objective function for response generation.

\subsection{Dialog State Tracking} Motivated by \citeauthor{lin-etal-2020-mintl} (\citeyear{lin-etal-2020-mintl}), we model the Levenshtein dialog state, which  means the difference between the current dialog state and the previous dialog
state, for dialog state tracking task to  generate minimal dialog state and reduce the inference latency.
The Levenshtein dialog state $\Delta{B_t}$ of dialog turn $t$,  is generated based on the previous dialog state $B_{t-1}$, the previous system response $R_{t-1}$, and the current user utterance $U_t$ via the encoder-decoder framework:

{\footnotesize\begin{equation}
H_{eb} = encoder(B_{t-1},R_{t-1},U_t),\\
\end{equation}%
\begin{equation}
\Delta{B_t} = decoder_b(H_{eb}),\\
\end{equation}}%
where $H_{eb}$ denotes the hidden representation of the encoder for dialog state tracking. Therefore, the dialog state tracking objective function can be optimized  by minimizing:

{\footnotesize
\begin{equation}
\mathcal{L}_{B} = \sum_{i=1}^{N} \sum_{t=1}^{n_i} -log P(\Delta{B_t}|B_{t-1},R_{t-1},U_t),
\end{equation}}%
where $N$ denotes the number of dialog sessions, $n_i$ denotes the number of dialog turns in the dialog session $i$.

For inference,  
a predefined function $\Omega (\cdot)$ is used to generate the dialog state $B_t$ as

{\footnotesize
\begin{equation}
B_t = \Omega (\Delta{B_t}, B_{t-1}).
\end{equation}}%
The predefined function $\Omega (\cdot)$  deletes the slot-value pair in $B_{t-1}$ when the \texttt{NULL} symbol appears in the $\Delta{B_t}$, and it updates the $B_{t-1}$ when new slot-value pair or new value for one slot appears in the $\Delta{B_t}$.
Refer to  \citeauthor{lin-etal-2020-mintl} (\citeyear{lin-etal-2020-mintl}) for more details. The generated dialog state $B_t$ is used to query the corresponding database. The database state embedding $DB_t$  represents the number of matched entities and whether the booking is available or not. The embedding $DB_t$ is used as the start token embedding of the response decoder for response generation.

\subsection{Response Generation} 

The response ${R_t}$ of dialog turn $t$ is generated based on the previous system response $R_{t-1}$, the current user utterance $U_t$,  the current dialog state $B_{t}$, and the database state embedding $DB_t$, which is formulated as:

{\footnotesize
\begin{equation}
H_{er} = encoder(R_{t-1},U_t,B_{t}),\\
\end{equation}%
\begin{equation}
R_t = decoder_r(H_{er},DB_t),\\
\end{equation}}%
where $H_{er}$ denotes the hidden representation of the encoder for response generation. Therefore, the response generation objective function can be optimized  by minimizing:

{\footnotesize
\begin{equation}
\mathcal{L}_{R} = \sum_{i=1}^{N} \sum_{t=1}^{n_i} -log P(R_t|R_{t-1},U_t,B_{t},DB_t).
\end{equation}}%


\section{Methodology}
In this section, we proposes two reconstruction strategies, i.e., back reconstruction and denoising reconstruction, respectively. 
Generally, during task-oriented dialog training,  objective functions $\mathcal{L}_{BR}$ and  $\mathcal{L}_{DR}$ are added to enhance model learning ability. The general  objective function of a task-oriented dialog system can be reformulated as follows:

{\footnotesize\begin{equation}
\begin{aligned}
\mathcal{L}_{all} = \mathcal{L}_{B} + \mathcal{L}_{R} + \lambda_{1} \mathcal{L}_{BR} + \lambda_{2} \mathcal{L}_{DR},  \label{loss}
\end{aligned}
\end{equation}}%
where  $\mathcal{L}_{BR}$ and $\mathcal{L}_{DR}$ denote the objective functions  for  back reconstruction and denoising reconstruction. $\lambda_{1}$ and $\lambda_{2}$ are hyper-parameters that adjust the weights of the objective functions.
\subsection{Back Reconstruction}

Dialog state is essential for the task completion of a task-oriented dialog system~\cite{LUNA}. 
 As illustrated in Figure~\ref{fig:BR}, we propose a back reconstruction strategy to mitigate the generation of inaccurate  dialog states, including encoder-reconstructor and encoder-decoder-reconstructor modules.
For the encoder-reconstructor module, the dialog context $C(t) = (B_{t-1},R_{t-1},U_t)$ could be reconstructed   to enhance encoder information representation by the encoder hidden representation $H_{eb}$.
For the encoder-decoder-reconstructor module, the decoder hidden representation $H_{db}$ could be used to reconstruct the dialog context $C(t)$ to encourage the dialog state decoder to achieve complete information of dialog context.

The dialog state would be reconstructed  back to the source
input and  the corresponding reconstruction
score would be calculated  to measure the adequacy of the dialog state.
The objective function $\mathcal{L}_{BR}$ for the back reconstruction is optimized by minimizing:

{\footnotesize
\begin{equation}
\begin{aligned}
\mathcal{L}_{BR}&= \sum_{i=1}^{N} \sum_{t=1}^{n_i} -log P(C(t)|H_{eb} )\\
&+ \sum_{i=1}^{N} \sum_{t=1}^{n_i} -log P(C(t)|H_{db} ).
\end{aligned}
\end{equation}}%
\subsection{Denoising Reconstruction}
To enhance the denoising capability of the task-oriented dialog system, we propose denoising reconstruction to guarantee  that the system  learns enough  dialog context representation  to recover the original version, as illustrated in Figure \ref{fig:dr}.
Motivated by denoising auto-encoder strategy that maps a corrupted input back to the original version~\cite{DBLP:journals/jmlr/VincentLLBM10}, we introduce noise in the form of random token deleting and masking in the source input  to improve the dialog model learning ability.
Specifically, we delete or mask every token in the previous dialog state and  system response with a probability $\alpha$. 
More concretely, we propose two denoising reconstruction modules, i.e., dialog state denoising  and response denoising modules.

For the dialog state denoising module, we reconstruct   the new Levenshtein dialog state, which  means the corrupted part of the dialog state rather than the complete  dialog state in the original denoising auto-encoder.
The Levenshtein dialog state $\Delta{B'_{t-1}}$ of dialog turn $t$, is generated based on the noisy dialog context $N_B(t) = (N(B_{t-1}),R_{t-1},U_t)$. $N(B_{t-1})$ is the previous corrupted dialog state.
For example, the Levenshtein dialog state `\textit{taxi\_destination=stevenage train station}' is reconstructed from the corrupted dialog state where `\textit{taxi\_destination}' is masked and `\textit{train}' is deleted,
as shown in Figure \ref{fig:dr}.
For response denoising module, the previous system response $R_{t-1}$ of dialog turn $t$ is reconstructed based on  the noisy dialog context $N_R(t) = (N(R_{t-1}),U_t,B_{t},DB_{t-1})$. $N(R_{t-1})$ is the previous noisy system response.
Therefore, the objective function $\mathcal{L}_{DR}$ for the denoising reconstruction is optimized by minimizing:

{\footnotesize
\begin{equation}
\begin{aligned}
\mathcal{L}_{DR} &= \sum_{i=1}^{N} \sum_{t=1}^{n_i} -log P(\Delta{B'_{t-1}}|N_B(t))\\
&+\sum_{i=1}^{N} \sum_{t=1}^{n_i} -log P(R_{t-1}|N_R(t)).
\end{aligned}
\end{equation}}%
\subsection{Training and Inference Details} 
There exists inconsistency between the lexicalized user utterance and delexicalized system response, which is used to reduce the influence  of different slot values on evaluation~\cite{DBLP:conf/aaai/ZhangOY20}. This adds an extra burden for the system to generate a delexicalized system response.
To alleviate this issue, we  introduce delexicalized user utterances for response generation  while lexicalized user utterances are still used for dialog state tracking. For example,  `\textit{02:15}' is converted into delexicalized form  `\textit{[taxi\_arriveby]}' for response generation, as shown in Figure~\ref{fig:baseline}.
Different forms of user utterances take  better training of both tasks, ultimately improving task completion.

 For inference of dialog state tracking,  generated previous dialog state, oracle previous system response, and  current user utterance are used as  dialog context to generate the current Levenshtein dialog state.
 For inference of response generation, motivated by~\citeauthor{DBLP:conf/aaai/YangLQ21} (\citeyear{DBLP:conf/aaai/YangLQ21}), we use generated previous system response instead of oracle previous system response to generate the current system response  to maintain coherence throughout  the whole dialog session.

\section{Experiments}

\begin{table*}[ht]
  \centering
  \scalebox{.92}{
	\begin{tabular}{llllll}
		\toprule
		\bf Model  &\bf Pre-trained &\bf Inform & \bf Success & \bf BLEU & \bf Combined \\ 
		\midrule
		DAMD~\cite{DBLP:conf/aaai/ZhangOY20} &n/a&76.3 &60.4 & 16.6 & 85.0\\
		SimpleTOD~\cite{DBLP:conf/nips/Hosseini-AslMWY20} &DistilGPT2&84.4&70.1&15.0&92.3\\
		MinTL-T5-small~\cite{lin-etal-2020-mintl}&T5-small&80.0&72.7& \bf19.1&95.5 \\
		SOLOIST~\cite{peng2020soloist} &GPT-2& 85.5&72.9&16.5&  95.7\\
		MinTL-BART~\cite{lin-etal-2020-mintl}&BART-large&84.9&74.9&17.9&97.8\\
		LAVA~\cite{lubis-etal-2020-lava}&n/a&91.8&81.8&12.0&98.8\\
		UBAR$^\ast$~\cite{DBLP:conf/aaai/YangLQ21}   &DistilGPT2&91.5&77.4&17.0&101.5\\
					SUMBT+LaRL~\cite{DBLP:journals/corr/abs-2009-10447}&BERT-base&92.2&85.4&17.9& 106.7  \\
        \cdashline{1-6}[1pt/2pt]
        Baseline (mask=0)  &T5-small& 89.0& 78.8& 17.9&101.8\\
        
        Baseline (mask=0.15)  &T5-small& 88.0& 77.6& 17.7&100.5\\

		   
		BORT& T5-small&\bf 93.8++& \bf 85.8++ &18.5&  \bf 108.3++ \\


		\bottomrule
	\end{tabular}}
	\caption{Comparison of end-to-end models evaluated on MultiWOZ 2.0. ``++" after a score indicates that the proposed BORT is significantly better than Baseline (mask=0) at significance level $p<$0.01. 
	$^\ast$ denotes the re-evaluated result by the author-released model since the result reported in this original paper~\cite{DBLP:conf/aaai/YangLQ21} was evaluated using the ground truth dialog state instead of generated dialog state to query the database entities. 
	\label{tab:e2emain_result}}
\end{table*}

\subsection{Datasets and Evaluation Metrics}
To establish our proposed end-to-end task-oriented dialog system, we consider two task-oriented dialog datasets, MultiWOZ 2.0~\cite{budzianowski-etal-2018-multiwoz} and  CamRest676~\cite{wen-etal-2017-network}.


MultiWOZ 2.0 is a large-scale human-to-human multi-domain task-oriented dialog dataset. The dataset consists of seven domains: attraction, hospital, police, hotel, restaurant, taxi, and train. It contains 8438, 1000, and 1000 dialog sessions for training, validation, and testing datasets. Each dialog session  covers 1 to 3 domains, and multiple different domains might be mentioned in a single dialog turn. 
Particularly, there are no hospital and police domains in the validation and testing dataset.
To make our experiments comparable with previous work \cite{DBLP:conf/aaai/ZhangOY20,lin-etal-2020-mintl,DBLP:conf/aaai/YangLQ21}, we use the pre-processing script released by \citeauthor{DBLP:conf/aaai/ZhangOY20} (\citeyear{DBLP:conf/aaai/ZhangOY20}) and follow the automatic evaluation metrics to evaluate the response quality for the task-oriented dialog system. \textbf{Inform}  rate measures if a dialog system has provided a correct entity; \textbf{Success} rate measures if a dialog system has provided a correct entity and answered all the requested information; \textbf{BLEU} score~\cite{papineni-etal-2002-bleu} measures  the fluency of the generated response; the \textbf{combined score}, which is computed by $ (Inform + Success) \times 0.5 + BLEU$, measures the overall quality of the dialog system. 
To evaluate the performance of dialog state tracking, 
we use the \textbf{joint goal accuracy} to measure the accuracy of generated dialog states.


CamRest676 is a small-scale restaurant-domain  dataset. It contains 408, 136, 132 dialog sessions for training, validation, and testing datasets.
To make our experiments comparable with previous work \cite{lei-etal-2018-sequicity,wu-etal-2021-alternating},
we use the same delexicalization strategy and use \textbf{BLEU} score and \textbf{Success F1} to evaluate the response quality for the task-oriented dialog system. The  success rate
measures if the system answered all the requested
information to assess  recall while Success F1 balances recall and precision.
\subsection{Settings}

In the training process for the task-oriented dialog system, we select two backbone models. BORT is a transformer-based system initialized by a pre-trained model. BORT\_G is a GRU-based system without a pre-trained model. The  detailed training settings and results of the BORT\_G backbone are  provided in  Appendix~\ref{moresettings}.

  For the BORT backbone, we use pre-trained T5-small~\cite{2020t5} to initialize the dialog system, based on the HuggingFace Transformers library~\cite{wolf-etal-2020-transformers} and follow the settings of \citeauthor{lin-etal-2020-mintl} (\citeyear{lin-etal-2020-mintl}). There are six layers for the encoder and the decoder. The dimension of hidden layers is set to 512, and the head of attention is 8. The batch size is set to 96. The AdamW optimizer~\cite{DBLP:conf/iclr/LoshchilovH19} is used to optimize the model parameters. The learning rate is 0.0025, and the learning rate decay is 0.8.
  The hyper-parameters $\lambda_{1}$ and $\lambda_{2}$ are set to 0.05 and 0.03, respectively. For the denoising reconstruction strategy, the noise probability $\alpha$ is 0.15. The hyper-parameter  analysis is provided in Appendix~\ref{sec:parameter}. Training early stops when no improvement on the combined score of the validation set for five epochs. All results  in the low-resource scenario are the average scores of  three runs. One P40 GPU is used to train all  dialog systems.

\subsection{Baselines}
Compared with other previous work, 
our proposed BORT is evaluated
 in two context-to-response settings:   end-to-end
modeling to generate dialog state and system  response, and policy optimization to generate system  response
based on ground truth dialog state. Policy optimization results are provided in Appendix~\ref{sec:po}.

Sequicity~\cite{lei-etal-2018-sequicity} and
DAMD~\cite{DBLP:conf/aaai/ZhangOY20} are  GRU-based end-to-end task-oriented dialog systems with a copy
mechanism. 
Decoder based pre-trained model GPT-2~\cite{radford2019language} is used in  SimpleTOD~\cite{DBLP:conf/nips/Hosseini-AslMWY20}, SOLOIST~\cite{peng2020soloist}, and UBAR~\cite{DBLP:conf/aaai/YangLQ21}. 
Encoder-decoder based pre-trained model T5~\cite{2020t5} and BART~\cite{lewis-etal-2020-bart} is used in MinTL~\cite{lin-etal-2020-mintl}. 
Reinforcement learning is used in LAVA~\cite{lubis-etal-2020-lava} and SUMBT+LaRL~\cite{DBLP:journals/corr/abs-2009-10447}. Especially, SUMBT+LaRL merges a dialog state tracking model SUMBT~\cite{lee-etal-2019-sumbt} and a dialog policy model LaRL~\cite{zhao-etal-2019-rethinking} and fine-tune them via reinforcement learning, achieving the state-of-the-art performance.  

In addition, we implement two baseline systems. One baseline is the base architecture of a task-oriented dialog system, as illustrated in Figure~\ref{fig:baseline}. The other is a noise-based  baseline system, just masking 15\% tokens in the dialog context for dialog training.

\subsection{Main Results}



Table~\ref{tab:e2emain_result} presents the detailed inform rates, success rates, BLEU scores, and combined scores of  end-to-end dialog models on the MultiWOZ 2.0. 
Our re-implemented baseline system performs better than MinTL~\cite{lin-etal-2020-mintl}, using the same pre-trained T5-small model. This indicates that the baseline is a strong system.  
Our proposed BORT significantly outperforms our re-implemented baseline system  by  6.5 combined scores, while the simple noise-based method (Baseline (mask=0.15)) doesn't achieve better performance.  
Moreover, BORT outperforms the previous state-of-the-art SUMBT+LaRL by 1.6 combined scores, achieving the best performance in terms of inform rate, success rate, and combined score. This demonstrates the effectiveness of our proposed BORT. 

\begin{table}[ht]
  \centering
  \scalebox{.82}{
	\begin{tabular}{lcc}
		\toprule
		\bf Model  & \bf Success F1& \bf BLEU \\ 
		\midrule
Sequicity~\cite{lei-etal-2018-sequicity}&85.4& 25.3\\
		ARDM~\cite{wu-etal-2021-alternating}&86.2&25.4\\
		SOLOIST~\cite{peng2020soloist} & 87.1 &25.5\\
        \cdashline{1-3}[1pt/2pt]
		BORT &  \bf 89.7& \bf 25.9\\

		\bottomrule
	\end{tabular}}\caption{Comparison of end-to-end  task-oriented dialog systems on CamRest676.\label{tab:camrest}}
\end{table}

To better assess the generalization capability of BORT, we fine-tune BORT on the CamRest676.
The  detailed Success F1 and BLEU scores on  the CamRest676  are  presented  in  Table~\ref{tab:camrest}.  Our proposed BORT outperforms the previous state-of-the-art SOLOIST by 2.6 Success F1, achieving the best performance in terms of Success F1. This demonstrates the generalization capability of our proposed BORT.

\subsection{Further Evaluation  Analysis}

\citeauthor{nekvinda-dusek-2021-shades} (\citeyear{nekvinda-dusek-2021-shades}) identify inconsistencies between previous task-oriented dialog methods in data preprocessing and evaluation metrics and introduce a standalone standardized evaluation script.
BLEU score is computed with references, which have been obtained from the delexicalized MultiWOZ 2.2 span annotations.

\begin{table}[ht]
  \centering
  \scalebox{.6}{
	\begin{tabular}{lcccc}
		\toprule
		\bf Model & \bf Inform & \bf Success & \bf BLEU & \bf Combined\\ 
		\midrule
		DAMD~\cite{DBLP:conf/aaai/ZhangOY20} &57.9 &47.6 & 16.4 &69.2\\
				LABES~\cite{zhang-etal-2020-probabilistic}&68.5&58.1&18.9&82.2\\
				AuGPT~\cite{DBLP:journals/corr/abs-2102-05126}&76.6&60.5&16.8&85.4\\
		MinTL-T5-small~\cite{lin-etal-2020-mintl}&73.7&65.4&\bf19.4 &89.0\\
		
		SOLOIST~\cite{peng2020soloist} & 82.3&72.4&13.6&91.0\\
		
		DoTS~\cite{DBLP:journals/corr/abs-2103-06648}&80.4&68.7&16.8&91.4\\
		UBAR~\cite{DBLP:conf/aaai/YangLQ21}   &  83.4&70.3&17.6 & 94.5\\
        \cdashline{1-5}[1pt/2pt]
		BORT & \bf 85.5 & \bf 77.4 & 17.9 & \bf 99.4 \\

		\bottomrule
	\end{tabular}}\caption{Comparison of end-to-end task-oriented dialog systems evaluated on the standardized  setting~\cite{nekvinda-dusek-2021-shades}.\label{tab:standardized_setting}}
\end{table}

To get a more complete picture of the effectiveness of  reconstruction strategies, we also use this evaluation script to evaluate our proposed BORT which is trained on MultiWOZ 2.0. As shown in Table~\ref{tab:standardized_setting}, BORT also substantially outperforms the previous state-of-the-art UBAR by a large margin (4.9 combined scores), achieving the best performance in terms of inform rate, success rate, and combined score. This further demonstrates the effectiveness of our proposed BORT.

\subsection{Ablation Study}
We empirically investigate the performance of the different components of  BORT as shown in Table~\ref{tab:ablation_1}. Our introduced user utterance delexicalization strategy gains 1.9 combined scores, indicating the effectiveness of the  user utterance delexicalization strategy. Back reconstruction performs slightly better than denoising reconstruction by 1 combined score regarding the two proposed reconstruction strategies. Moreover, the combination of both reconstruction strategies can complement each other to further improve the performance of the  dialog system.
The detailed analysis on different modules of every reconstruction strategy is provided in   Appendix~\ref{AS_ap}.
\begin{table}[ht]
  \centering
  \scalebox{.65}{
	\begin{tabular}{lcccc}
		\toprule
		\bf Model & \bf Inform & \bf Success & \bf BLEU & \bf Combined\\ 
		\midrule

BORT&93.8 & 85.8 & 18.5 & 108.3\\
\;\;\;\;\;w/o DR & 92.9&84.0&18.8&107.3 \\
\;\;\;\;\;w/o BR&92.0&84.4&18.1&106.3\\
\;\;\;\;\;w/o BR \& DR& 90.4 & 81.4  & 17.8 & 103.7\\
\;\;\;\;\;w/o BR \& DR\& UD & 89.0&78.8&17.9&101.8\\
		\bottomrule
	\end{tabular}}\caption{The performance of the different components of our proposed BORT on MultiWOZ 2.0. BR denotes back reconstruction, DR denotes denoising reconstruction, UD denotes user utterance delexicalization.\label{tab:ablation_1}}
\end{table}

\subsection{Dialog State Tracking}
\label{DST}

Table~\ref{tab:dst} reports the dialog state tracking performance of the end-to-end task-oriented dialog systems on MultiWOZ 2.0.  BORT substantially outperforms MinTL~\cite{lin-etal-2020-mintl} using the same pre-trained T5-small model  by  2.8 points, achieving 54.0 joint goal accuracy. Moreover, BORT achieves the highest joint goal accuracy among the end-to-end task-oriented dialog systems. This indicates that our proposed reconstruction strategies could improve dialog state learning ability.

\begin{table}[ht]
  \centering
  \scalebox{.82}{
	\begin{tabular}{lc}
		\toprule
		\bf Model & \bf Joint Accuracy \\ 
		\midrule
		MinTL-T5-small~\cite{lin-etal-2020-mintl}&51.2\\
		SUMBT+LaRL~\cite{DBLP:journals/corr/abs-2009-10447}&51.5\\
		
		MinTL-BART~\cite{lin-etal-2020-mintl}&52.1\\
		UBAR~\cite{DBLP:conf/aaai/YangLQ21}   & 52.6\\
		SOLOIST~\cite{peng2020soloist} &53.2\\
		\cdashline{1-2}[1pt/2pt]
BORT& \bf 54.0 \\
		\bottomrule
	\end{tabular}}\caption{The dialog state tracking performance of  end-to-end task-oriented dialog systems on MultiWOZ 2.0.\label{tab:dst}}
\end{table}

\subsection{Case Study and Human Evaluation}

Moreover, we  analyze translation examples and conduct  a human evaluation to further explore  the effectiveness of BORT. 
Figure \ref{fig:case} shows an example  generated by MinTL and BORT, respectively. More examples are provided in  Appendix~\ref{More examples}.
MinTL generates the response to request for the preferred area about college since it generates  an inaccurate dialog state `\textit{attraction\_type=college}' rather than correct dialog state `\textit{attraction\_name=jesus college}'. In contrast, BORT generates an accurate dialog state, achieving the  appropriate response that provides the information of \textit{jesus college}. These further demonstrate the effectiveness of our proposed reconstruction strategies.
\begin{figure}[ht]
	\centering
	\includegraphics[width=0.45\textwidth]{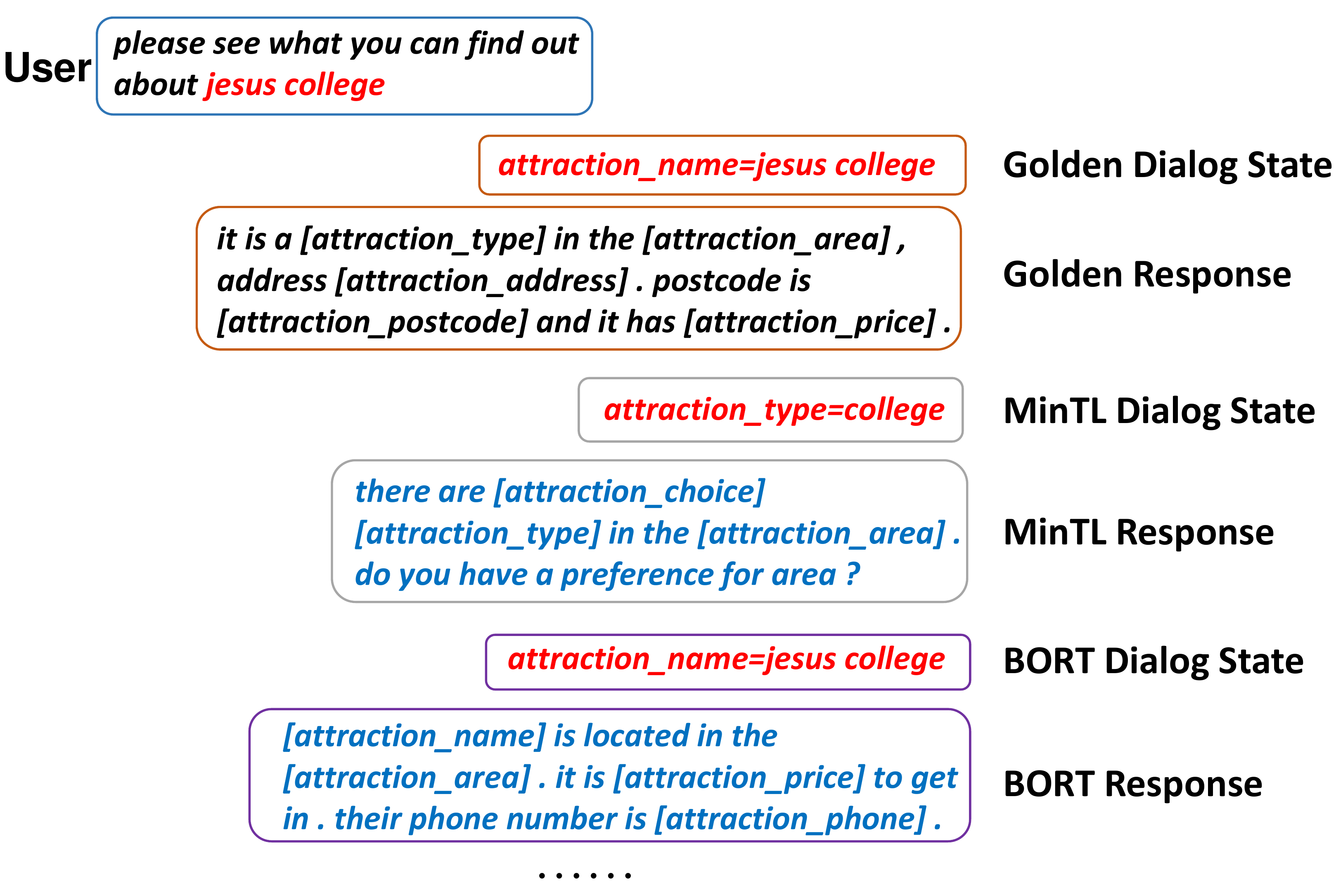}
	\caption{ An example of the task-oriented dialog systems  in dialog session PMUL4025. \label{fig:case}} 
\end{figure}

For human evaluation, we manually evaluate the quality of generated responses on 50 dialog sessions, which are randomly extracted from the MultiWOZ 2.0 testing set. We consider the fluency and appropriateness of the generated response based on scores ranging from 1  to 5. 
The fluency metric measures whether the generated response is fluent. The appropriateness metric measures whether the generated response is appropriate and the system understands the user's goal.
Three fluent English speakers are asked to evaluate these generated responses.
The average scores evaluated by them are shown in Table~\ref{tab:human}. The results are consistent with the automatic evaluation, indicating  that  BORT could improve the quality of generated response. 
\begin{table}[h]
  \centering
  \scalebox{.82}{
	\begin{tabular}{lcc}
		\toprule
		\bf Model & \bf  Fluency  &  \bf Appropriateness\\ 
		\midrule
		MinTL-T5-small& 4.50&3.88\\
		UBAR   & 4.50&3.81\\
BORT& \bf4.55 &\bf3.98 \\
		\bottomrule
	\end{tabular}}\caption{The human evaluation of the end-to-end task-oriented dialog systems on MultiWOZ 2.0.\label{tab:human}}
\end{table}
\subsection{Domain Adaptation Analysis}
To investigate  the domain adaptation ability of BORT to generalize to some unseen domains, we simulate zero-shot  experiments by excluding one domain  and training BORT on other domains. As shown in Table~\ref{tab:domain}, the train and taxi domains achieve the highest combined scores because they have  a high overlap in ontology with other domains. In addition,  BORT and MinTL with an encoder-decoder-based pre-trained model  achieve significantly better domain adaptation performance than  DAMD without a pre-trained model and  UBAR with a decoder-based pre-trained model, which demonstrates the encoder-decoder based pre-trained model have better domain transferability. Moreover, our proposed reconstruction strategy could further improve combined scores in the zero-shot domain scenario.

\begin{table}[ht]
   \small
   \centering
   \setlength{\tabcolsep}{1mm}{\scalebox{.93}{
	\begin{tabular}{lcccccccccc}
		\toprule
		\multirow{.9}*{\bf Model} &\multicolumn{1}{c}{\bf Attraction } &   \multicolumn{1}{c}{\bf Hotel}&   \multicolumn{1}{c}{\bf Restaurant}&\multicolumn{1}{c}{\bf Taxi }&\multicolumn{1}{c}{\bf Train }\\ 
		\midrule
DAMD&28.7&26.9&24.4&52.3&51.4\\
UBAR&28.3&29.5&23.5&59.5&53.9\\
MinTL&33.4&37.3&31.5&60.4&77.1\\
BORT&\bf 33.6&\bf38.7&\bf32.0&\bf 62.7&\bf85.6\\

		\bottomrule
	\end{tabular}}\caption{Comparison of combined scores on the MultiWOZ 2.0 in the zero-shot domain scenario.\label{tab:domain}}
	}

\end{table}

\begin{table*}[ht]
   \small
   \centering
   \scalebox{.62}{
	\begin{tabular}{lcccccccccccccccc}
		\toprule
		\multirow{2}*{\bf Model} &\multicolumn{4}{c}{\bf 5\% } &\multicolumn{4}{c}{\bf 10\% } &   \multicolumn{4}{c}{\bf 20\%}&  \multicolumn{4}{c}{\bf 30\% }\\ 
\cmidrule(lr){2-5} \cmidrule(lr){6-9} \cmidrule(lr){10-13} \cmidrule(lr){14-17} 
		& \bf Inform  & \bf Success & \bf BLEU & \bf Combined& \bf Inform  & \bf Success & \bf BLEU& \bf Combined& \bf Inform  & \bf Success & \bf BLEU& \bf Combined& \bf Inform  & \bf Success & \bf BLEU& \bf Combined\\ 
		\midrule
DAMD&49.1&23.7&11.3&47.7& 57.6   &   32.6  &   12.0 & 57.1   &64.7&45.0&15.3&70.2&64.5&47.3&15.5&71.4\\
UBAR&35.7&21.2&11.0&39.5& 62.4   &   43.6  &  12.7  &  65.7  &76.2&58.3&14.1&81.4&81.2&65.4&14.7&88.0 \\
MinTL&55.2&40.9&\bf13.9&62.0&  67.7  &  55.7   & 15.3   & 77.0   &66.7&57.9&\bf17.3&79.6&74.9&66.5&\bf17.3&88.0\\
BORT&\bf69.8&\bf45.9&11.0&\bf68.9&  \bf 74.5 & \bf60.6    & \bf 15.5  & \bf 83.1  &\bf82.1&\bf65.5&14.3&\bf88.1&\bf83.8&\bf69.9&17.2&\bf94.1\\

		\bottomrule
	\end{tabular}}\caption{Comparison of task-oriented dialog systems on the MultiWOZ 2.0 in the low resource scenarios.\label{tab:low_resource}}

\end{table*}
\subsection{Low Resource Scenario Analysis}
To better assess the robustness of our proposed BORT, we choose 5\%, 10\%, 20\%, and 30\% of training dialog sessions to investigate the performance of task-oriented dialog systems in the  low resource scenario. As shown in Table~\ref{tab:low_resource},  BORT substantially outperforms other methods  in these low-resource scenarios. This is because the error propagation problem in the low resource scenario is more serious, while  BORT could effectively alleviate the error propagation problem.
Moreover, our proposed BORT trained on the 30\% dataset performs  comparably to some baseline systems trained on all datasets as shown in Table~\ref{tab:e2emain_result}. These further demonstrate that our proposed BORT is robust, alleviating poor performance in the low-resource scenario.

\subsection{Error Propagation Analysis}
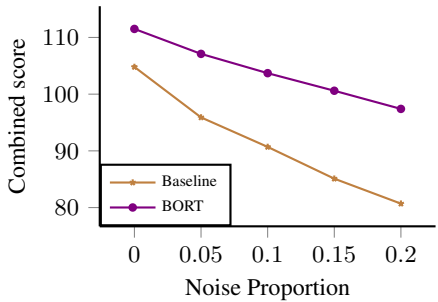
\begin{figure}[h]
\setlength{\abovecaptionskip}{0pt}
\begin{center}
\pgfplotsset{height=4.5cm,width=8.5cm,compat=1.15,every axis/.append style={thick}}
\begin{tikzpicture}
\tikzset{every node}=[font=\small]
\begin{axis}
[width=6cm,enlargelimits=0.13, tick align=outside, legend style={cells={anchor=west},legend pos=south west, legend columns=1,every axis legend/.append style={at={(0,0.)}}}, xticklabels={ $0$, $0.05$,$0.1$, $0.15$, $0.2$},
xtick={0,1,2,3,4},
		axis y line*=left,
		axis x line*=left,
ylabel={Combined score},xlabel={Noise Proportion},font=\small]

\addplot+ [sharp plot, mark=star,mark size=1.2pt,mark options={mark color=brown}, color=brown] coordinates
{ (0,104.8)(1,95.9) (2,90.7) (3,85.1) (4,80.7)};
\addlegendentry{\tiny Baseline}

\addplot+ [sharp plot, mark=*,mark size=1.2pt,mark options={mark color=violet}, color=violet] coordinates
{(0,111.5) (1,107.1)(2,103.7) (3,100.6) (4,97.4)};
\addlegendentry{\tiny BORT}

\end{axis}
\end{tikzpicture}
\caption{\label{fig:error_propagation}The policy optimization performance (combined score) of baseline  and BORT as the noise in the oracle dialog state increases on the MultiWOZ 2.0.}

\end{center}
\end{figure}

To investigate  the denoising capability of our proposed BORT, we perform the simulated experiments, where noise is added in the  oracle dialog state for the policy optimization evaluation. In detail, we replace every token in the  oracle dialog state with the masked token with a probability  to generate synthetic noise.
 As shown in Figure \ref{fig:error_propagation},   BORT performs substantially better than the baseline system in the noisy scenario. 
 In particular, as the noise ratio in the oracle dialog state increases, the performance gap between the baseline system and BORT increases. 
 When noise proportion is 0, BORT still performs better than the baseline system because BORT generates more appropriate response via the denoising reconstruction strategy.
 These demonstrate that our proposed BORT is robust and effective.

\section{Related Work}

End-to-end  task-oriented dialog system has attracted much attention in the dialog community. 
A two-stage copynet framework was proposed to establish an end-to-end task-oriented dialog system based on a single sequence-to-sequence model
\cite{lei-etal-2018-sequicity}.
\citeauthor{DBLP:conf/aaai/ZhangOY20} (\citeyear{DBLP:conf/aaai/ZhangOY20}) proposed a multi-action data augmentation framework to improve the diversity of dialog responses.  Recently, large-scale language model pre-training has been  effective for enhancing  many natural language processing tasks~\cite{peters-etal-2018-deep,radford2018improving,devlin-etal-2019-bert}. Decoder-based pre-trained language model such as GPT-2~\cite{radford2019language} was used to improve the performance of end-to-end task-oriented dialog system~\cite{budzianowski-vulic-2019-hello,DBLP:conf/nips/Hosseini-AslMWY20,peng2020soloist,DBLP:conf/aaai/YangLQ21}.
The Levenshtein dialog  state instead of the dialog state was generated to reduce the inference latency~\cite{lin-etal-2020-mintl}. In addition, they used encoder-decoder-based pre-trained model such as T5~\cite{2020t5} and BART~\cite{lewis-etal-2020-bart} to establish a dialog system. In contrast with previous work, in which system response was generated, \citeauthor{wu-etal-2020-tod} (\citeyear{wu-etal-2020-tod}) used encoder-based pre-trained model such as BERT~\cite{devlin-etal-2019-bert} for task-oriented dialog system, aiming
to retrieve the most relative system response from
a candidate pool.
Reinforcement learning  could also be used to enable task-oriented dialog systems to achieve more successful task completion~\cite{lubis-etal-2020-lava,DBLP:journals/corr/abs-2009-10447}.

\citeauthor{DBLP:conf/aaai/TuLSLL17} (\citeyear{DBLP:conf/aaai/TuLSLL17}) proposed an encoder-decoder-reconstructor framework for neural machine translation to alleviate over-translation and
under-translation problems. Reconstruction strategy was used  to moderate
dropped pronoun translation problems~\cite{DBLP:conf/aaai/WangTSZGL18}. In contrast, we considered the adequacy of semantic representations rather than natural language sentences to build the reconstruction model.
\citeauthor{DBLP:journals/jmlr/VincentLLBM10} (\citeyear{DBLP:journals/jmlr/VincentLLBM10}) proposed a denoising autoencoder, in which random noise is added to enhance the robustness of the model,  alleviating the overfitting problem of traditional auto-encoder.
The denoising auto-encoder strategy was used as the language model to generate more fluent translation candidates for the unsupervised neural machine translation~\cite{DBLP:conf/iclr/ArtetxeLAC18,lample-etal-2018-phrase,sun-etal-2019-unsupervised}. In addition, 
a denoising auto-encoder was used to  pre-train sequence-to-sequence models on the large scale corpus~\cite{lewis-etal-2020-bart,liu-etal-2020-multilingual-denoising}. In contrast, we proposed a denoising reconstruction mechanism to alleviate the error   propagation problem   along   the   multi-turn conversation flow.
\section{Conclusion}
This paper proposes back and denoising reconstruction strategies for the end-to-end task-oriented dialog system. Back reconstruction strategy is proposed to   mitigate   the   generation   of inaccurate   dialog  states, achieving better task completion of the task-oriented dialog system. Denoising reconstruction is used to train a robust task-oriented dialog system, further alleviating  the error  propagation problem.
  Our extensive experiments and analysis on MultiWOZ 2.0 and CamRest676 demonstrate the effectiveness of our proposed  reconstruction strategies.
\section*{Acknowledgments}
We are grateful to the anonymous reviewers and the area chair for their insightful comments and suggestions.
This work is supported by the National Key Research and Development Program of China under Grant No. 2020AAA0108600.

\bibliography{custom}
\bibliographystyle{acl_natbib}

\clearpage
\appendix
\section{Appendix}
\label{sec:appendix}
\subsection{BORT\_G Settings and Results}
\label{moresettings}

 For the BORT\_G backbone, we follow the settings of \citeauthor{DBLP:conf/aaai/ZhangOY20} (\citeyear{DBLP:conf/aaai/ZhangOY20}). We use one layer bi-directional GRU for the encoder and the decoder. The dimension of hidden layers is set to 100. The batch size is 128. The AdamW optimizer~\cite{DBLP:conf/iclr/LoshchilovH19} is used to optimize the model parameters, and the learning rate is 0.005.
 
 
 \begin{table}[ht]
  \centering
  \scalebox{.75}{
	\begin{tabular}{lllll}
		\toprule
		\bf Model  &\bf Inform & \bf Success & \bf BLEU & \bf Combined \\ 
		\midrule
		DAMD &76.3 &60.4 & 16.6 & 85.0\\
		MinTL-T5-small&80.0&72.7& 19.1&95.5 \\
        \cdashline{1-5}[1pt/2pt]

		   
		BORT\_G &87.3 & 75.8 & 18.4 &  100.0 \\

		\bottomrule
	\end{tabular}}
	\caption{Comparison of end-to-end  models evaluated on MultiWOZ 2.0.
	\label{tab:damdmain_result}}
\end{table}
 
 As shown in Table ~\ref{tab:damdmain_result}, BORT\_G  performs better than DAMD without a pre-trained  model,  achieving the improvement of  15.0 combined scores, even though multi-action data augmentation is not used in BORT\_G. Moreover, BORT\_G outperforms MinTL~\cite{lin-etal-2020-mintl}, using the pre-trained  model. This demonstrates the effectiveness and applicability of our proposed  reconstruction strategies. 
\subsection{Hyper-parameter Analysis}
\label{sec:parameter}

\begin{figure}[h]
\setlength{\abovecaptionskip}{0pt}
\begin{center}
\pgfplotsset{height=5.6cm,width=8.5cm,compat=1.15,every axis/.append style={thick}}
\begin{tikzpicture}
\tikzset{every node}=[font=\small]
\begin{axis}
[width=7cm,enlargelimits=0.13, tick align=outside, legend style={cells={anchor=west},legend pos=south west, legend columns=3,every axis legend/.append style={at={(0,0.95)}}}, xticklabels={ $0$, $0.01$,$0.03$, $0.05$, $0.1$,$0.3$, $0.5$},
xtick={0,1,2,3,4,5,6},
		axis y line*=left,
		axis x line*=left,
ylabel={Combined score},xlabel={$\lambda_{1}$},font=\small]

\addplot+ [sharp plot, mark=square*,mark size=1.2pt,mark options={mark color=cyan}, color=cyan] coordinates
{ (0,100.5) (1,104.1) (2,104.4) (3,105.1) (4,106.0) (5,104.0)(6,102.9)};
\addlegendentry{\tiny $\lambda_{2}=0$}

\addplot+ [sharp plot, mark=diamond*,mark size=1.2pt,mark options={mark color=pink}, color=pink] coordinates
{ (0,103.8) (1,104.9) (2,106.0) (3,107.4)(4,106.2)(5,104.3)(6,101.2)};
\addlegendentry{\tiny $\lambda_{2}=0.01$}

\addplot+ [sharp plot, mark=*,mark size=1.2pt,mark options={mark color=violet}, color=violet] coordinates
{(0,105.2) (1,105.8)(2,107.2) (3,109.7) (4,106.1)(5,103.2)(6,101.9) };
\addlegendentry{\tiny $\lambda_{2}=0.03$}

\addplot+ [sharp plot, mark=star,mark size=1.2pt,mark options={mark color=brown}, color=brown] coordinates
{ (0,106.7)(1,108) (2,105.7) (3,104.2) (4,104)(5,102.5)(6,102)};
\addlegendentry{\tiny $\lambda_{2}=0.05$}

\addplot+ [sharp plot, mark=pentagon*,mark size=1.2pt,mark options={mark color=red}, color=red] coordinates
{ (0,105.6)(1,105.6) (2,106) (3,104.8) (4,103.6)(5,102.4)(6,102)};
\addlegendentry{\tiny $\lambda_{2}=0.1$}

\addplot+ [sharp plot, solid,mark=Mercedes star,mark size=1.2pt,mark options={mark color=green}, color=green] coordinates
{ (0,97.1)(1,97.4) (2,97.6) (3,94.0) (4,93.3)(5,93.1)(6,94.0)};
\addlegendentry{\tiny $\lambda_{2}=0.3$}

\addplot+ [sharp plot,solid, mark=x,mark size=1.2pt,mark options={mark color=yellow}, color=yellow] coordinates
{ (0,89.9)(1,94.3) (2,94.1) (3,94.9) (4,95.1)(5,96.8)(6,93.8)};
\addlegendentry{\tiny $\lambda_{2}=0.5$}

\addplot+ [sharp plot,densely dashed,no markers, color=gray] coordinates
{ (0,100.5) (1,100.5) (2,100.5) (3,100.5) (4,100.5) (5,100.5)(6,100.5)};
\addlegendentry{\tiny Baseline}
\end{axis}
\end{tikzpicture}
\caption{\label{fig:lambda}BORT performance  (combined score) with different levels of hyper-parameters on the MultiWOZ 2.0 validation set.}

\end{center}
\end{figure}
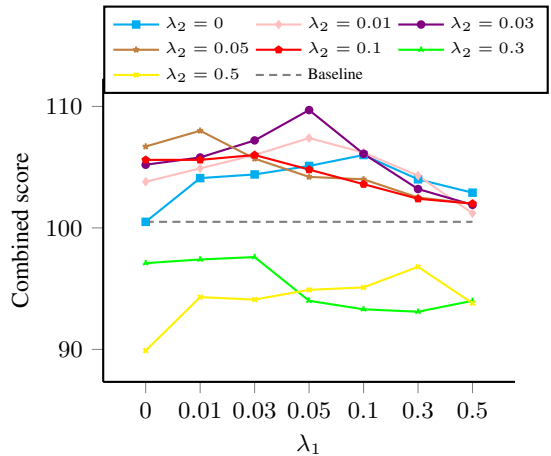

In Figure \ref{fig:lambda}, we empirically investigate how the hyper-parameters  in Eq. \ref{loss} affects the dialog performance on the MultiWOZ 2.0 validation set. The selection of hyper-parameters  $\lambda_{1}$ and $\lambda_{2}$ influence the role of the $\mathcal{L}_{BR}$ and $\mathcal{L}_{DR}$ across the entire end-to-end task-oriented dialog training process.
Larger values of $\lambda_{1}$ or $\lambda_{2}$ cause the $\mathcal{L}_{BR}$ or $\mathcal{L}_{DR}$ to play a more important role than the original task-oriented dialog loss terms. The smaller the value of $\lambda_{1}$ or $\lambda_{2}$, the less important is the $\mathcal{L}_{BR}$ or $\mathcal{L}_{DR}$. As   Figure \ref{fig:lambda} shows, $\lambda_1$ ranging from 0.01 to 0.5 nearly all enhances task-oriented dialog performance, and  when $\lambda_2$ is larger than 0.3, the performance  underperforms the baseline system.
When  $\lambda_1 =0.05$ and $\lambda_2 =0.03$, our proposed BORT achieves the best performance on the validation set.

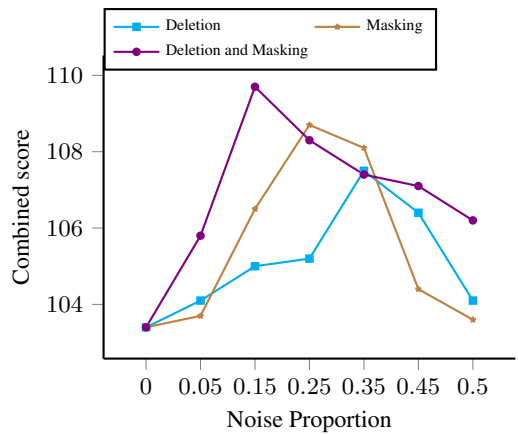
\begin{figure}[h]
\setlength{\abovecaptionskip}{0pt}
\begin{center}
\pgfplotsset{height=5.6cm,width=8.5cm,compat=1.15,every axis/.append style={thick}}
\begin{tikzpicture}
\tikzset{every node}=[font=\small]
\begin{axis}
[width=7cm,enlargelimits=0.13, tick align=outside, legend style={cells={anchor=west},legend pos=south west, legend columns=2,every axis legend/.append style={at={(0,0.95)}}}, xticklabels={ $0$, $0.05$,$0.15$, $0.25$, $0.35$,$0.45$, $0.5$},
xtick={0,1,2,3,4,5,6},
		axis y line*=left,
		axis x line*=left,
ylabel={Combined score},xlabel={Noise Proportion},font=\small]

\addplot+ [sharp plot, mark=square*,mark size=1.2pt,mark options={mark color=cyan}, color=cyan] coordinates
{ (0,103.4) (1,104.1) (2,105.0) (3,105.2) (4,107.5) (5,106.4)(6,104.1)};
\addlegendentry{\tiny Deletion}

\addplot+ [sharp plot, mark=star,mark size=1.2pt,mark options={mark color=brown}, color=brown] coordinates
{ (0,103.4)(1,103.7) (2,106.5) (3,108.7) (4,108.1)(5,104.4)(6,103.6)};
\addlegendentry{\tiny Masking}

\addplot+ [sharp plot, mark=*,mark size=1.2pt,mark options={mark color=violet}, color=violet] coordinates
{(0,103.4) (1,105.8)(2,109.7) (3,108.3) (4,107.4)(5,107.1)(6,106.2) };
\addlegendentry{\tiny Deletion and Masking}

\end{axis}
\end{tikzpicture}
\caption{\label{fig:noise}BORT performance  (combined score) with different levels of noise type and noise proportion on the MultiWOZ 2.0 validation set.}

\end{center}
\end{figure}

In addition, the influence of noise type and noise proportion on the performance of our proposed BORT  on the MultiWOZ 2.0 validation set is empirically investigated, as shown in Figure~\ref{fig:noise}. Both   deletion and masking noise strategies could improve the dialog performance. In particular, their combination is further better than both of them.  This demonstrates that both noise strategies can complement each other to further improve the dialog performance. As shown in Figure~\ref{fig:noise}, when the noise proportion is 0.15, our proposed BORT achieves the best performance on the validation set.

\subsection{ Policy Optimization Evaluation}
\label{sec:po}
\begin{table*}[ht]
  \centering
  \scalebox{.81}{
	\begin{tabular}{llllll}
		\toprule
		\bf Model  &\bf Pre-trained  &\bf Inform & \bf Success & \bf BLEU & \bf Combined \\ 
		\midrule

		LaRL~\cite{zhao-etal-2019-rethinking}&n/a&82.8&79.2&12.8   &93.8\\
		SimpleTOD~\cite{DBLP:conf/nips/Hosseini-AslMWY20} &DistilGPT2&88.9&67.1&16.9&94.9\\
		HDSA~\cite{chen-etal-2019-semantically} &BERT-base&82.9&68.9&\bf23.6 & 99.5 \\
		ARDM~\cite{wu-etal-2021-alternating} &GPT-2&87.4&72.8&20.6 & 100.7\\
		DAMD~\cite{DBLP:conf/aaai/ZhangOY20} &n/a&89.2&77.9&18.6&102.2\\
		
		SOLOIST~\cite{peng2020soloist} &GPT-2&89.6&79.3&18.0&102.5\\
	
		UBAR~\cite{DBLP:conf/aaai/YangLQ21}   &DistilGPT2&94.0&83.6&17.2&106.0  \\ 
	
		LAVA~\cite{lubis-etal-2020-lava}&n/a&\bf 97.5& \bf94.8&12.1&108.3  \\
	HDNO~\cite{DBLP:conf/iclr/WangZKG21}   &n/a&96.4&84.7&18.9&109.5  \\
        \cdashline{1-6}[1pt/2pt]

		BORT\_G &n/a&89.6&80.5&19.1& 104.2\\
		BORT &T5-small&  96.1 & 88.8 & 19.0 & \bf 111.5\\

		\bottomrule
	\end{tabular}}
	\caption{Comparison of policy optimization models evaluated on MultiWOZ 2.0. \label{tab:PL_main_result}}
\end{table*}
 The detailed inform rates, success rates, BLEU scores, and combined scores of  policy optimization dialog models on the MultiWOZ 2.0 are presented in Table~\ref{tab:PL_main_result}.
The ground truth dialog state is used for the policy optimization setting to query the database entities and generate system responses.
Our proposed BORT achieves performance comparable to the state-of-the-art LAVA in terms of inform rate. In addition, compared with previous policy optimization methods,  BORT 
achieves better performance in terms of the combined score even though BORT has not modeled action learning.

Compared with previous works, BORT achieves much more significant improvement in the end-to-end setting rather than policy optimization setting because our proposed reconstruction strategies pay more attention to improving the quality of dialog state while the golden dialog state is used in the policy optimization setting. 

\subsection{Ablation Study}
\label{AS_ap}
Moreover, we further investigate the performance of the different components of the two proposed reconstruction strategies, respectively. As shown in Table~\ref{tab:ablation_2},   encoder-decoder-reconstructor module for  back reconstruction strategy significantly outperforms   encoder-reconstructor module by 2.2 combined scores because dialog state decoder could achieve more dialog context information for encoder-decoder-reconstructor. In addition,  regarding  two denoising reconstruction modules, dialog state denoising  and response denoising  have achieved similar performance. These two modules could improve the denoising capability of the task-oriented dialog system.
\begin{table}[ht]
  \centering
  \scalebox{.65}{
	\begin{tabular}{lcccc}
		\toprule
		\bf Model & \bf Inform & \bf Success & \bf BLEU & \bf Combined\\ 
		\midrule
Back reconstruction &92.9&84.0&18.8&107.3 \\
\; \; w/o enc-rec&92.2&83.5&19.0&106.9 \\
\; \; w/o enc-dec-rec&92.1&81.2&18.0&104.7 \\
	\midrule
Denoising reconstruction&92.0&84.4&18.1&106.3\\
\; \; w/o dialog state denoising &91.7&83.0&17.9&105.3\\
\; \; w/o response denoising &92.8&81.2&18.6&105.6\\

		\bottomrule
	\end{tabular}}\caption{The performance of the different components of the two proposed reconstruction strategies. enc-dec denotes encoder-reconstructor module, enc-dec-rec denotes encoder-decoder-reconstructor module.\label{tab:ablation_2}}
\end{table}

\subsection{More examples}
\label{More examples}
\begin{figure}[ht]
	\centering
	\includegraphics[width=0.45\textwidth]{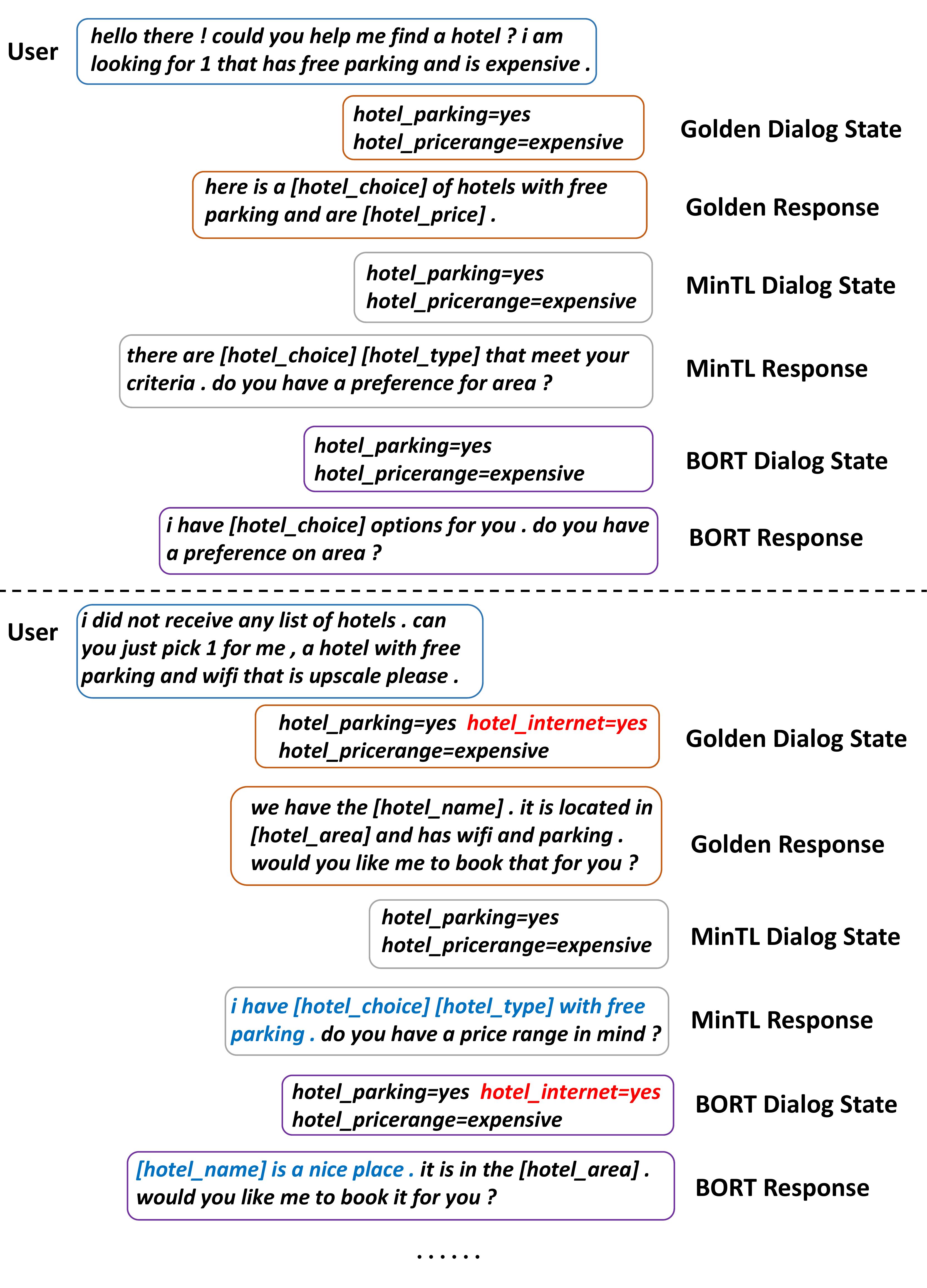}
	\caption{ An example of the task-oriented dialog systems  in dialog session MUL1139. \label{fig:case5}} 
\end{figure}

\begin{figure}[ht]
	\centering
	\includegraphics[width=0.45\textwidth]{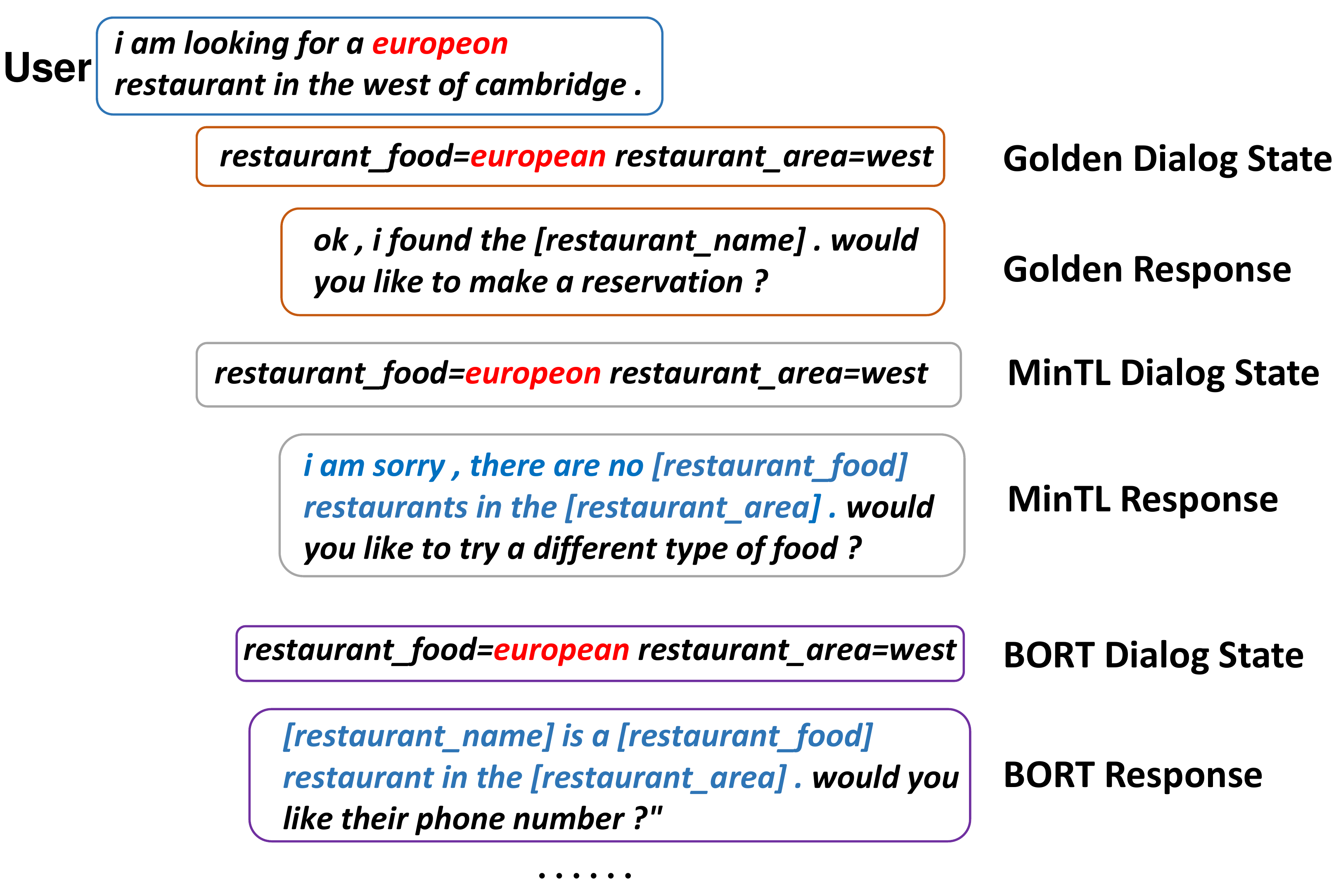}
	\caption{ An example of the task-oriented dialog systems  in dialog session PMUL0095. \label{fig:case1}} 
\end{figure}

\begin{figure}[ht]
	\centering
	\includegraphics[width=0.45\textwidth]{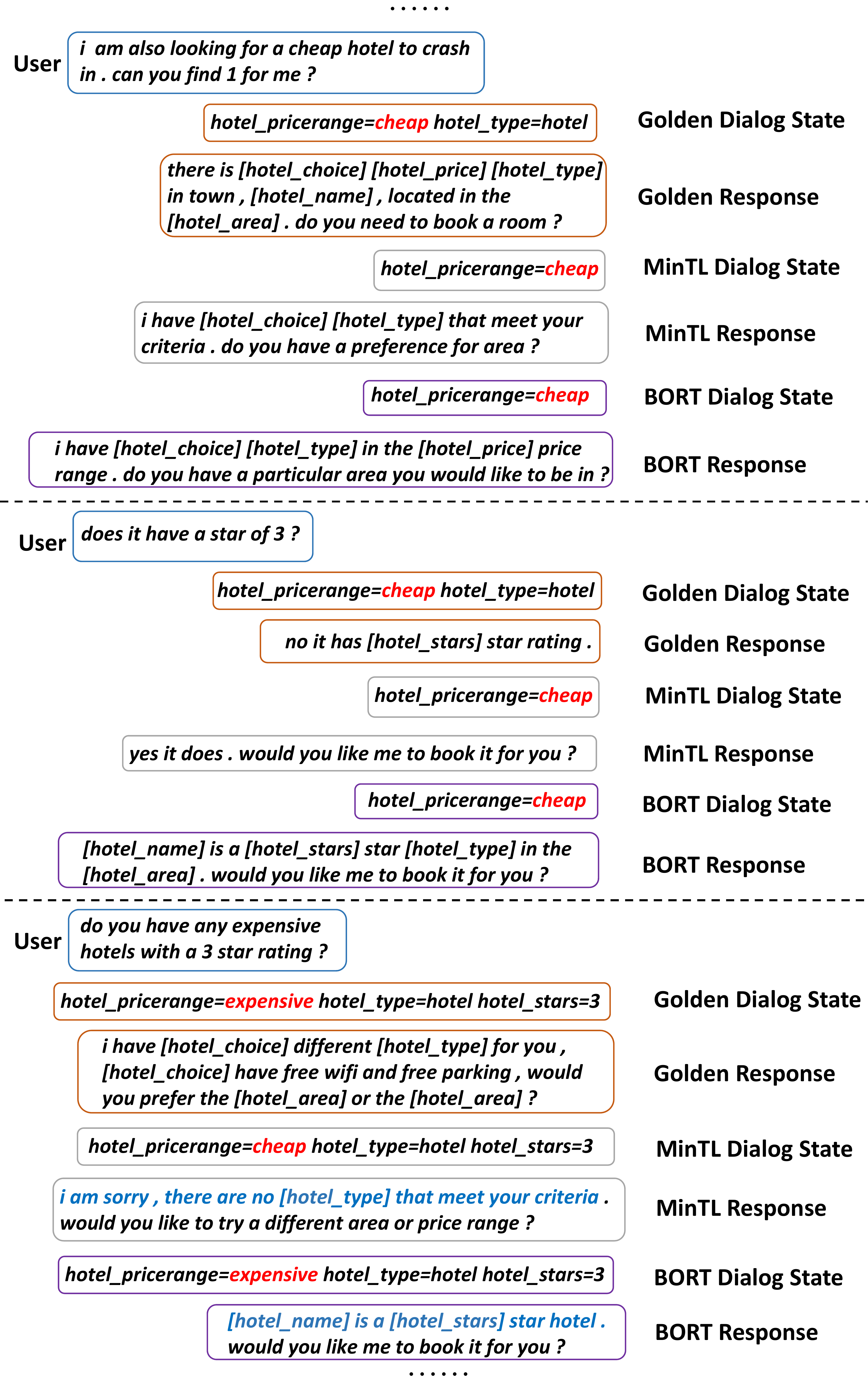}
	\caption{ An example of the task-oriented dialog systems  in dialog session PMUL3868. \label{fig:case2}} 
\end{figure}

\begin{figure}[ht]
	\centering
	\includegraphics[width=0.45\textwidth]{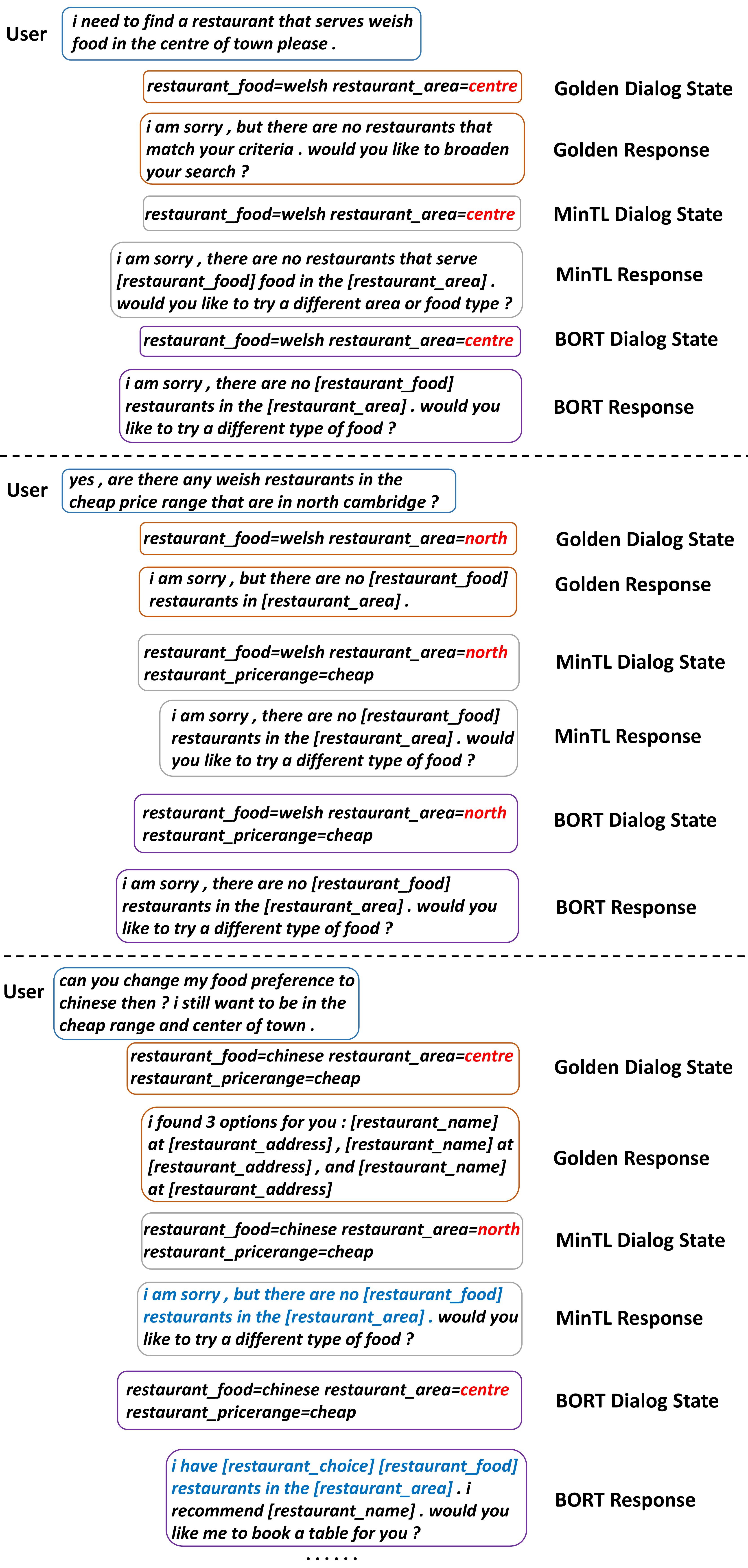}
	\caption{ An example of the task-oriented dialog systems  in dialog session MUL0286. \label{fig:case3}} 
\end{figure}

\begin{figure}[ht]
	\centering
	\includegraphics[width=0.45\textwidth]{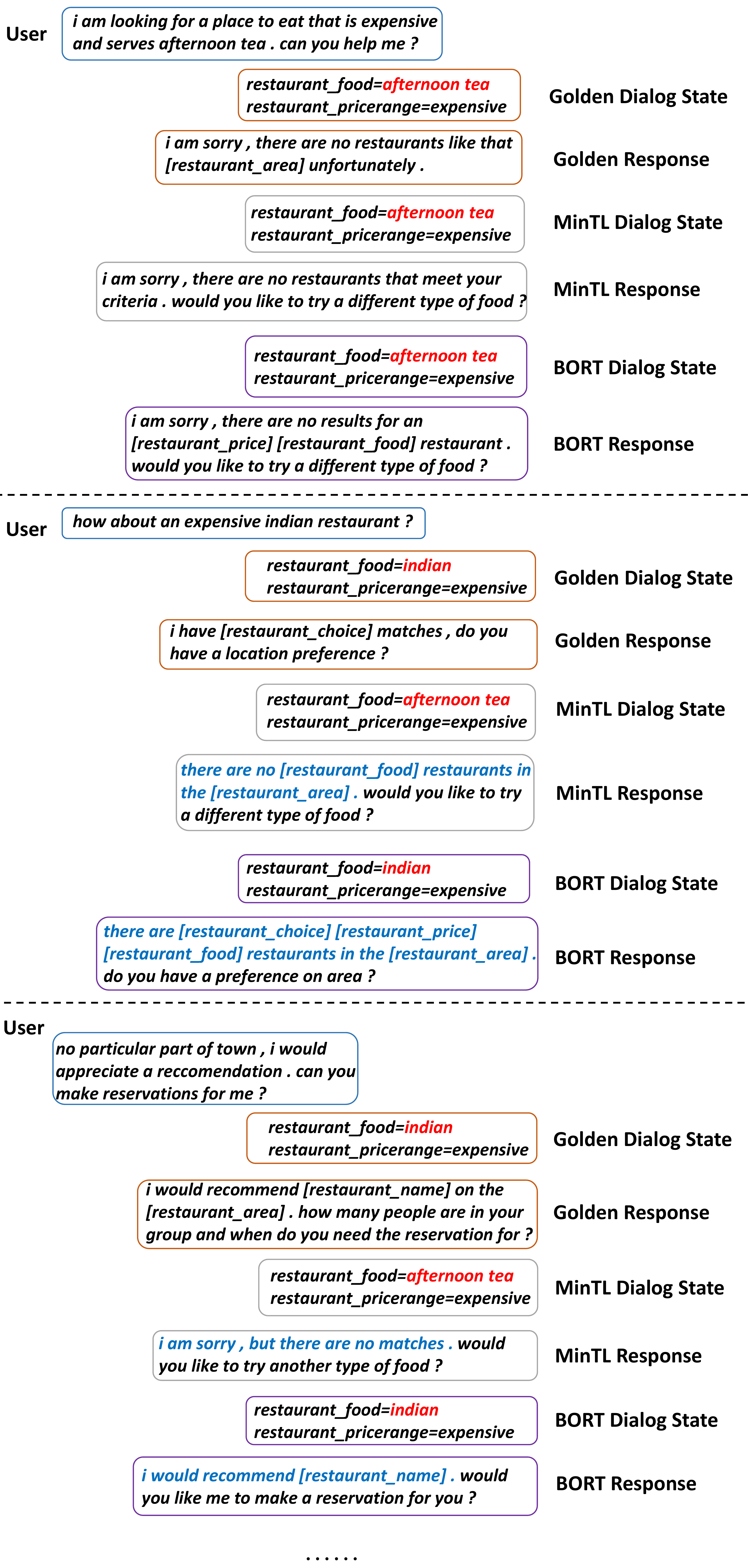}
	\caption{ An example of the task-oriented dialog systems  in dialog session PMUL3875. \label{fig:case4}} 
\end{figure}

Figures \ref{fig:case5} - \ref{fig:case4} show several examples generated by MinTL and BORT, respectively. 
As shown in Figure \ref{fig:case5},  MinTL generates the inadequate dialog state, which may provide the hotel without internet. Our proposed BORT  reconstructs the generated dialog state back to the original input context to ensure the information in the input side is completely transformed to the output side to achieve an adequate dialog state via a back reconstruction strategy.
 Figure \ref{fig:case1} shows that our proposed BORT generated the correct slot value '\textit{european}' rather than the corrupted one '\textit{europeon}' from the corrupted dialog context, indicating the robustness of  the denoising reconstruction strategy.  As shown in Figures \ref{fig:case2} -  \ref{fig:case4}, MinTL generates the inaccurate dialog state, leading to the inaccurate response. The results are consistent with our opinion that the generated dialog state, which is crucial for task completion of a task-oriented dialog system, has always been inaccurate across the end-to-end task-oriented dialog system training. Moreover, Figure \ref{fig:case4} shows that MinTL faces  the  problem  of  error  propagation from both previously generated inaccurate dialog states and responses. Our proposed BORT  can alleviate these issues via reconstruction strategies, further demonstrating the effectiveness of BORT.


\end{document}